\documentclass[letterpaper, 10 pt, conference]{ieeeconf}  

\IEEEoverridecommandlockouts                              

\usepackage{fancyhdr}
\newcommand{\mytitle}{\textbf{Accepted final version.}
To appear in \textit{Proc. of the IEEE Conference on Decision and Control, 2025}.\\
\copyright 2025 IEEE. Personal use of this material is permitted. Permission
from IEEE must be obtained for all other uses, in any current or future
media, including reprinting/republishing this material for advertising or
promotional purposes, creating new collective works, for resale or
redistribution to servers or lists, or reuse of any copyrighted component of
this work in other works.}
\usepackage{graphics} 
\usepackage{amsmath} 
\usepackage{amssymb}  
\usepackage{algorithm}
\usepackage[noend]{algorithmic}
\usepackage{bbm}
\usepackage{cite}
\usepackage{comment}
\usepackage{pgfplots}
\usepackage{caption}
\usepackage{subcaption} 
\usepackage{float}
\usepackage{balance}
\usepackage{hyperref}
\usepackage{url}

\usepackage{tikz}

\newtheorem{assumption}{Assumption}
\newtheorem{definition}{Definition}
\newtheorem{proposition}{Proposition}

\newcommand{\St}{\mathcal{S}}
\newcommand{\A}{\mathcal{A}}

\newcommand{\R}{\mathbb{R}}
\newcommand{\EE}{\mathbb{E}}
\newcommand{\MG}{\mathcal{G}}
\DeclareMathOperator{\sgn}{sgn}

\newcommand{\fakepar}[1]{\vspace{1mm}\noindent\textbf{#1.}}
\newcommand{\capt}[1]{\mdseries{\emph{#1}}}

\title{\LARGE \bf
Beyond expected value: geometric mean optimization\\ for long-term policy performance in reinforcement learning
}

\author{Xinyi Sheng$^{1}$ and Dominik Baumann$^{1}$
\thanks{*We acknowledge the financial support of the Finnish Ministry of Education and Culture through the Intelligent Work Machines Doctoral Education Pilot Program (IWM VN/3137/2024-OKM-4)}
\thanks{$^{1}$Cyber-physical Systems Group, Aalto University, Espoo, Finland
        {\tt\small \{xinyi.sheng, dominik.baumann\}@aalto.fi}}%
}

\begin{document}

\maketitle
\thispagestyle{empty}
\pagestyle{empty}

\begin{abstract}
Reinforcement learning (RL) algorithms typically optimize the expected cumulative reward, i.e., the expected value of the sum of scalar rewards an agent receives over the course of a trajectory. 
The expected value averages the performance over an infinite number of trajectories. However, when deploying the agent in the real world, this ensemble average may be uninformative for the performance of individual trajectories.
Thus, in many applications, optimizing the long-term performance of individual trajectories might be more desirable. 
In this work, we propose a novel RL algorithm that combines the standard ensemble average with the time-average growth rate, a measure for the long-term performance of individual trajectories.
We first define the Bellman operator for the time-average growth rate. We then show that, under multiplicative reward dynamics, the geometric mean aligns with the time-average growth rate. To address more general and unknown reward dynamics, we propose a modified geometric mean with $N$-sliding window that captures the path-dependency as an estimator for the time-average growth rate. This estimator is embedded as a regularizer into the objective, forming a practical algorithm and enabling the policy to benefit from ensemble average and time-average simultaneously. 
We evaluate our algorithm in challenging simulations, where it outperforms conventional RL methods.
\end{abstract}

\section{INTRODUCTION}
Reinforcement learning (RL) is a popular method for learning control policies from data when no dynamics model is available and has shown great success in various fields~\cite{mnih2015human,duan2016benchmarking}.
However, many breakthrough results were achieved in simulated or gaming environments.
Transferring those results to the real world remains challenging due to the non-robust nature of many RL algorithms~\cite{russell2015research,amodei2016concrete,uesato2018rigorous}.
RL algorithms typically try to find optimal policies by optimizing the expected value of a reward signal they receive when trying out actions, as recommended by standard textbooks~\cite{sutton2018reinforcement,bertsekas2019reinforcement}.
The expected value averages over infinitely many rollouts of a particular policy.
In environments with heavy-tailed rewards, optimizing this average can yield risky policies that achieve exceptionally high rewards in a few cases but fail in all others.
In control, we are often interested in policies that work well for the entire lifetime of, for instance, a mobile robot.
Thus, such extremely risky policies are not desirable.

Instead, we are interested in optimizing the long-term performance of a single trajectory, which is captured by the notion of the \emph{time-average} reward.
Optimizing the time-average instead of the expected reward can be achieved through a suitable reward transformation~\cite{baumann2023reinforcement}.
Nevertheless, this requires us to collect a reward trajectory first, learn a transformation, and then use the transformed data to optimize our policy.
Thus, this basically restricts us to offline or episodic algorithms.
In this paper, we take a different route.
We first show how, for particular reward dynamics, replacing the expected value with the geometric mean allows us to optimize the time-average growth rate. The specific reward dynamics follow a geometric Brownian motion (GBM) as a tractable model to estimate the time-average growth rate. 
We then modify the geometric mean to adapt it to the general dynamics assumed in RL, introduce a sliding window approach to estimate the time-average growth rate, and propose a convex combination of the geometric mean and the expected value to yield an adequate optimization objective for unknown reward dynamics.
Importantly, our approach remains grounded in the standard additive RL framework, using the geometric mean as a regularizer. 

\fakepar{Contributions}
We make the following contributions:
\begin{itemize}
    \item We show that the geometric mean estimates the time-average growth rate in multiplicative reward dynamics;
    \item We derive the Bellman operator for optimizing the time-average growth rate of an unknown reward function;
    \item We propose a novel RL algorithm that optimizes a convex combination of a modified version of the geometric mean and the expected value;
    \item We evaluate our algorithm in challenging simulations.
\end{itemize}










\section{RELATED WORK}

In this section, we relate our contributions to the literature.

\fakepar{Regularized Markov decision process} 
We incorporate the geometric mean as an additional term in the RL optimization objective, inspired by regularized Markov decision processes (MDPs). 
The idea of regularized MDPs was proposed in \cite{todorov2006linearly} using the Kullback-Leibler (KL) divergence, and extended as a general framework in \cite{geist2019theory} based on \cite{scherrer2015approximate}. Since then, many variants have emerged \cite{vieillard2020munchausen,abdolmaleki2018maximum,vieillard2020leverage} with different regularization terms, improving the performance and robustness of RL methods. However, most of them focus on processing policies, but rarely capture the reward dynamics. Our approach introduces a reward-level regularization term based on a modified geometric mean, which is orthogonal and complementary to existing regularized MDPs and can be naturally integrated into these frameworks.

\fakepar{Multi-step learning} When considering the time average of an individual trajectory, multi-step learning is a natural approach. The idea of multi-step bootstrapping was introduced in \cite{peng1994incremental,de2018multi} to incorporate long-term rewards. Combining it with Q-learning has been further investigated in \cite{carvalho2023multi,yuan2019novel,hessel2018rainbow}. 
Those approaches often rely on a simple reward aggregation approach that cannot handle heavy-tailed rewards. 

\fakepar{Reward shaping} Using the geometric mean instead of the expected value can be interpreted as a form of reward shaping. 
Reward shaping has been explored in several works \cite{ng1999policy,memarian2021self,baumann2023reinforcement} to refine the reward signal.
However, only few studies have reshaped multi-step rewards. Similar to \cite{zheng2018learning}, we treat the time-average as an additional reward embedded in the objective function; however, whereas their method learns intrinsic rewards based on the expected value, our approach uses a more effective way to capture long-term rewards.

\fakepar{Time-average growth rate} Estimating the time-average growth rate in unknown environments is a challenging problem. The time-average growth rates for additive dynamic and multiplicative dynamic are derived in \cite{adamou2021microfoundations}. The existence of the time average growth rate for multiple dynamics has been shown in \cite{peters2018time}. Although~\cite{peters2018time} derives the time-average growth rate for given dynamics, it lacks practical algorithms that deal with unknown dynamics.

\fakepar{Average reward RL}
We aim to optimize the long-term performance of RL agents.
Average reward RL~\cite{zhang2021policy,wei2022provably,wang2023robust} tries the same by letting time in the reward function go to infinity.
However, they still take the expected value of the reward function.
Thus, with heavy-tailed rewards, we may learn similarly risky policies as traditional RL methods.

\section{PROBLEM SETTING}
We consider a standard Markov decision process (MDP) with tuple $\{ \St ,\A, P, r, \gamma \}$, where $\St $ is a finite state space, $\A$ is a finite action space, and $\gamma\in(0,1)$ is a discount factor. We denote $|X|$ as the cardinality of a set $X$ and $\Delta_{X}$ as the set of probability distributions over the set $X$. Then, $r:\St\times\A\to \R $ is the reward function and $P\in\Delta_{\St}^{|\St|\times|\A|}$ is the Markovian transition kernel, specifically $P(s'|s,a)$ denotes the probability of transiting to state $s'$ from $s$ when action $a$ is applied. Let $\tilde{R}_T = \sum_{t = 0}^{T-1} \gamma^t r_t$ and $R_T = \sum_{t = 0}^{T-1} r_t$  denote the discounted and undiscounted cumulative rewards over horizon $T$, respectively, where $r_t := r(S_t,A_t)$ is the per-step reward. In RL, the agent learns a policy $\pi\in\Delta_{\A}^{|\St|}$ by interacting with the environment, i.e., applying actions and observing states and rewards following from those actions. 

The general objective in RL is to find an optimal policy $\pi^*$ that can maximize the expected cumulative reward~\cite{sutton2018reinforcement}, 
\begin{align*}
    \pi^*=\arg\max_{\pi}\EE^\pi\left[ \sum_{t=0}^{\infty}\gamma^tr_t \right] = \arg\max_{\pi}\EE^\pi\left[\lim_{T\to\infty}\tilde{R}_T\right],
\end{align*}
where the expectation is taken over an ensemble of infinitely many trajectories. For the $i$-th trajectory we write  $\tilde{R}^{i}_T=\sum_{t=0}^{\infty}\gamma^tr_t$, so that 
\begin{align*}
    \EE[\tilde{R}_T] =\lim_{N\to\infty}\frac{1}{N} \sum_{i=1}^N R^{i}_T.
\end{align*}
In an ergodic system, the ensemble average coincides with the time-average return along any individual trajectory, i.e., $\lim_{T\to\infty}\frac{1}{T}R^{i}_T$, as shown in the ergodic theorem \cite{birkhoff1931proof}.
However, if the reward dynamic is non-ergodic, this equivalence breaks down, and the expected return may poorly represent what an agent experiences in a single trajectory. 
As an illustrative example, consider a coin-toss process with multiplicative reward dynamics and initial reward $R_0=100$. At each time step $t$, the agent receives $r_t = +0.5 R_t$ on heads or $r_t = -0.4 R_t$ on tails. While the expected cumulative reward grows exponentially, the time-average almost surely decays to zero. Such discrepancies are known to cause failures in state-of-the-art RL algorithms, as shown in~\cite{baumann2023reinforcement}, where the authors also discuss the relevance of non-ergodic reward dynamics in RL more generally.

Technically speaking, different policies lead to different cumulative reward dynamics. RL provides many approaches to learn policies $\pi$ such that the Q-value function
\begin{align}
    q^{\pi}(s_0,a_0) = \EE^\pi \left[\sum_{t=0}^\infty \gamma^t r_t | s_0,a_0\right]
\end{align}
is optimized based on the ensemble average. 
An effective way to resolve potential non-ergodicity in the reward dynamics is to find an ergodicity transformation $u$, which can convert non-ergodic reward dynamics to ergodic reward dynamics. After the transformation, the increment $r_t = R_{t+1} - R_t$ at each time step $t$ becomes $g_t = u(R_{t+1})-u(R_t)$, which is the \textit{growth rate at time $t$}. With an ergodicity transformation, optimizing the ensemble average also optimizes the time-average growth rate, that is
\begin{equation}
    \max_{\pi} \EE \left[ \lim_{T\to\infty} \frac{1}{T}\sum_{t = 0}^{T-1} g_t\right] \sim \max_{\pi} \lim_{T\to\infty} \frac{1}{T}\sum_{t=0}^{T-1}g_t, \label{equ: gr equal ens}
\end{equation}
where the left term is equivalent to $ \max_{\pi} \EE [\lim_{T\to\infty}u(R_T)] $ since $u(R_0)$ does not play any role in the optimization problem. 
The right term is the maximization problem through the lens of the time-average growth rate. Thus, if we can find a policy that maximizes not only the ensemble average but also the time average, the robustness of RL methods for individual trajectories can be improved.  

Nevertheless, it is challenging to find suitable ergodicity transformations in the real world, particularly in high-dimensional systems or unknown dynamical environments. When knowledge of multivariate real-world dynamics is lacking, it becomes necessary to develop more direct and general approaches. In this paper, we introduce a novel objective function that simultaneously captures both aspects by incorporating the long-term time-average objective as a regularization term into the conventional RL optimization criterion. We denote by $G^\pi_\infty (s,a) := \lim_{T\to\infty} \frac{1}{T}\sum_{t=0}^{T-1}g_t$ the time-average growth rate for each initial state-action pair $(s,a)$ under a given policy $\pi$. As time tends to infinity, the effect of stochasticity diminishes, and distinct policies yield different dynamical behaviors. We therefore propose the following new objective:
\begin{equation}
    \max_{\pi} \left\{ (1-\lambda) \EE^\pi\left[ \sum_{t=0}^\infty \gamma^t r_t  \right] + \lambda G^\pi_\infty  \right\},  \label{equ: objective}
\end{equation}
where $\lambda\in[0,1]$ is a balance parameter that determines the relative emphasis on the time-average growth rate compared to the conventional ensemble average reward. Rather than learning an explicit ergodicity transformation from the reward dynamics and applying it in the algorithm (``inside-out”), we estimate the time-average growth rate directly from observed rewards (``outside-in"). Although our method does not replace the ergodicity transformation in a formal sense, it achieves a similar effect by encouraging consistency between time-average and ensemble-average, thereby enabling path-dependent behavior to be captured without explicitly modeling the reward dynamics.



\section{AGGREGATION TECHNIQUES}
In this section, we propose a method for optimizing the objective~\eqref{equ: objective} by embedding the time-average growth rate as a regularization term into the standard RL objective. Notably, the reward dynamics vary with different policies, and under an appropriate ergodicity transformation, the optimal policy should be consistent with respect to both ensemble and time averages. When the dynamics are unknown and only rewards are available through interaction with the environment, we develop an estimation method---\textit{modified geometric mean over $N$-sliding window}---to facilitate the selection of an optimal policy based on the time average. This estimate is then incorporated into a regularized MDP framework. 

Our approach is built upon the following two assumptions:
\begin{assumption}
    The Markov chain underlying the MDP is irreducible and aperiodic. \label{Assp:1}
\end{assumption}
\begin{assumption}
    An ergodicity transformation exists for the reward dynamics. \label{Assp:2}
\end{assumption}
The first assumption ensures a stationary distribution for the MDP, as supported by \cite[Thm.~5.4]{levin2017markov}. The rationale of the second assumption is established in \cite{peters2018time}, suggesting that knowledge of the reward dynamics alone can determine the existence of such a transformation. Although our method does not explicitly utilize the ergodicity transformation, its theoretical existence is essential to justify our framework.

\subsection{Background}
\label{sec:background}
We now proceed to the foundations of our approach, starting with an overview of the regularized MDP framework. 

\subsubsection{Regularized Markov Decision Processes}
Following \cite{geist2019theory,vieillard2020leverage}, the standard Bellman operator is $ T^\pi q = r + \gamma P\langle\pi,q\rangle $ and $q^\pi$ is its unique fixed point, where $\langle \pi, q \rangle = (\langle\pi(\cdot|s), q(s,\cdot)\rangle)_{s\in\St}$ for any $q\in\R^{|\St|\times|\A|}$. With a slight abuse of notation, we let $\pi_s$ represent the vector $\pi(\cdot|s)$ and $q_s$ represent the vector $q(s,\cdot)$ when given state $s$.  In particular, $\langle\pi_s, q_s\rangle = \pi_s^Tq_s$ is the inner product when given a state $s$, and for each state we construct the vector $\langle \pi, q \rangle\in\R^{|S|}$. 

Then, in general, a strongly convex function $\Omega(\pi):\Delta_\A^{|\St|}\to \R^{|S|}$ is applied to a policy $\pi$ as a regularization term, which forms the regularized Bellman operator $ T^\pi_{\Omega}q = r + \gamma P( \langle\pi,q\rangle -\Omega(\pi) )$ and the regularized term is embedded in the optimization objective. 

Accordingly, in this paper, we apply the same scheme from regularized modified policy iteration (reg-MPI), consisting of a greedy step and an evaluation step,
\begin{equation}
    \left\{
    \begin{aligned}
    \pi_{t+1} &= \MG_{\Omega}(q_t), \\ 
    q_{t+1} &= (T^{\pi_{t+1}}_{\Omega})^m q_t.   
    \end{aligned}
    \right.  \label{equ: framework}
\end{equation}
In~\eqref{equ: framework}, the scheme starts with an initial policy $\pi_0$ and Q-function $q_0$. The greedy step generates the subsequent policy using a greedy operator defined as $ \MG_{\Omega} (q) = \arg\max_{\pi} ( \langle\pi,q\rangle -\Omega(\pi) )$. During the evaluation step, the Q-function is updated by applying the regularized Bellman operator $m$ times. In our work, we focus exclusively on deterministic policies, defined as $\pi(s) = \arg\max_{a\in\A} q(s,a)$. 

\subsubsection{Multi-step Q-learning}
For $m=1$, the scheme generates a regularized value iteration algorithm, whereas for $m\to\infty$, we obtain a regularized policy iteration algorithm. We typically use a multi-step rollout in the evaluation step to consider future rewards for each individual.  
In general, the $m$-step cumulative reward is defined as 
\begin{equation}
    \delta^{(m)}_t = \sum_{k=0}^{m-1} \gamma^{k}r_{t+k} . \label{equ:m-return}
\end{equation}
In multi-step Q-learning, the next Q-value $q_{t+1}(s_t,a_t)$ is updated by 
\begin{equation*}
    q_t(s_t,a_t) + \alpha ( \delta^{(m)}_t + \gamma^m \max_{a'} q(s_{t+m},a') - q_t(s_t,a_t)),
\end{equation*}
which has been used in \cite{hessel2018rainbow,yuan2019novel}.

\subsection{Time-Average Growth Rate Estimation}
Having introduced the required background, we turn to developing our own method.
To solve~\eqref{equ: objective}, we require an estimate of the time-average growth rate.
We propose two key estimation techniques for the time-average growth rate. First, we introduce an $N$-sliding window to manage the intractability of infinite horizon calculations. Second, we employ a modified geometric mean to effectively estimate the growth rate. These two techniques form the cornerstone of our approach, enabling a practical estimation of the time-average growth rate prior to applying our algorithm.

\subsubsection{Sliding window mechanism}
In non-ergodic reward dynamics, individual rollouts heavily depend on the path, and for a given $(s,a)$ pair, it is difficult to predict the corresponding reward without reward history. This inherent path-dependence contributes to the intractability of the problem. To address this challenge, we employ a sliding-window approach, which has been widely used to represent the complete historical reward trajectory~\cite{carvalho2023multi,cayci2024finite}. Specifically, we utilize a sliding window of fixed width $N$ to summarize the history by representing cumulative reward information over an interval of size $N$ as 
\begin{equation}
    \hat{R}_t = \sum_{i=0}^{N-1} r_{t+i}. \label{equ:reginal reward}
\end{equation}

\subsubsection{Modified geometric mean}
We now propose a modified geometric mean (MGM) to estimate the time-average growth rate. For tractability, we start by considering multiplicative dynamics, specifically, geometric Brownian motion (GBM), to model the reward dynamics. 
The time-average growth rate of GBM has been proved to be stable in \cite{peters2022ergodicity}. 
We will then generalize the approach to unknown reward dynamics.

To illustrate the cumulative reward dynamics, we assume that for each specific $S_t$, the reward increment $r_t=r(S_t) = R_{t} - R_{t-1}$ follows a GBM characterized by two parameters: the percentage drift $\mu(S_t)$ and the percentage volatility $\sigma(S_t)$. 
The overall reward dynamic is then described by the stochastic differential equation $ dR_t = \mu(S_t) R_t\, dt + \sigma(S_t) R_t\, dW_t $, where $ W_t $ is a standard Wiener Process. 

The following proposition establishes the existence of a fixed time-average growth rate in the above setting.
\begin{proposition} \label{prop:1}
    Let $i = 1, 2, ..., K $ denote the states in a finite state space $\St$ of size $K$. Under Assumption~\ref{Assp:1} and for a deterministic policy $\pi$, let $d^\pi_i$ denote the stationary distribution associated with state $i$, and let $\mu_i$ and $\sigma_i$ denote the percentage drift and the percentage volatility, respectively, corresponding to state $i$. Then, the cumulative reward dynamic possesses a fixed time-average growth rate 
    $$ \bar{g} = \sum_{i=1}^K d^\pi_i \left(\mu_i - \frac{\sigma_i^2}{2}\right). $$
\end{proposition}
All proofs are collected in the appendix. The proposition provides an exact solution for the time-average growth rate, which converges to a fixed incremental ratio $\bar{g}$ as $T$ goes to infinity. In the infinite-time limit, the noise term vanishes, yielding $R_{T+1} = \bar{g}R_T$. This multiplicative dynamic can be captured by the standard geometric mean 
\begin{align}
    \left(\frac{R_{T+\Delta t}}{R_T}\right)^{\frac{1}{\Delta t}}  = \bar{g}. \label{equ:normal GM}
\end{align}

Although the standard geometric mean theoretically provides a reasonable objective under ideal conditions, its application in practical scenarios is hindered by several factors. 
First, we only have finitely many samples that are subject to noise. Second, without considering historical information, estimating the time-average growth rate for the cumulative reward dynamic becomes intractable. 
Lastly, as the geometric mean involves taking roots, it cannot handle negative values.

To overcome these challenges, we introduce the MGM that leverages a sliding window of size $N$ to capture long-term reward features. 
The MGM is defined as
\begin{equation}
    \hat{G}_t = \sgn(\hat{R}_t)(|\hat{R}_t|)^{\frac{1}{N}}, \label{equ:modified GM}
\end{equation}
where $\hat{R}_t$ is the $N$-sliding window in~\eqref{equ:reginal reward} and $\sgn(\cdot)$ the sign function, ensuring that the estimator reflects the growth direction. 
To mitigate sensitivity issues associated with using a simple fraction in~\eqref{equ:normal GM}, we adopt the cumulative reward over an $N$-sliding window, since, for example, in~\eqref{equ:normal GM} the ratio of two negative values can yield a positive number, potentially misrepresenting a downward trend. The monotonicity of the cumulative reward smooths the fluctuations while simultaneously preserving the relative scale of the values. In the following, we leverage the MGM to estimate the time-average growth rate and integrate this estimator into our algorithm. 

\subsection{Learning algorithm}
We first assume an idealized setting where the time-average growth rate is perfectly known. Under this assumption, we define a Q-function corresponding to our objective in~\eqref{equ: objective} and derive the associated Bellman operator. We establish the key properties of the operator, which provide the theoretical guarantees for our proposed objective. This idealized framework forms the foundation of our approach. In subsequent sections, we extend this framework by invoking our MGM estimator for the time-average growth rate. This extension leads to a practical algorithm that can enhance the robustness of conventional methods.

\subsubsection{Theoretical properties}
In Section~\ref{sec:background}, we discussed that for the regularized MDP, we replace $\langle\pi,q\rangle$ by a regularized function $ \langle\pi,q\rangle -\Omega(\pi) $.
We propose to use the invariant time-average growth rate as the regularization term.
Then, we have $ \Omega(\pi) = \langle\pi,G^\pi_{\infty}\rangle \in\R^{\St}$ and $\Omega(\pi_s)= \langle\pi(\cdot|s),G^\pi_{\infty}(s,\cdot)\rangle$ as the state-wise term. Moving the balance parameter inside the expectation, the objective can be formally expressed in terms of the Q-function as
\begin{equation*}
    q^{\pi}_{\Omega}(s,a) = \EE^{\pi} \left[ \sum_{t=0}^\infty \gamma^t (1-\lambda)r_t + \lambda G^\pi_\infty \vert s_0 ,a_0\right],
\end{equation*}
where $G^\pi_\infty$ denotes the ideal time-average growth rate, which is constant for each state-action pair $(s,a)$, thereby acting as a linear coefficient for the regularization term $\Omega(\pi)$. Since our method focuses on deterministic policies, we now denote $q^{\pi}_{\Omega}$ as $q^{\pi}_{G}$. Next, we derive an $N$-step regularized optimality Bellman operator based on the traditional optimality Bellman operator in \cite{carvalho2023multi}. The traditional one is
\begin{align*}
    &T^N q (s_0,a_0) = \EE \left[r_0 + \gamma\max_{a_1}\EE\left[r(s_1,a_1) + \right.\right.\\
    & \left. \left.\gamma\max_{a_2}\EE\left[ r(s_2,a_2) +\cdots +\gamma\max_{a_N}q_(s_N,a_N)   \right]  \right]\vert s_0,a_0\right],
\end{align*}
where $q^\pi = \EE[\sum_{t=0}^\infty\gamma^tr_t] $ with fixed point $ \max_{\pi}q^\pi := q^*= T^N q^*$. Now, let $\pi^*$ represent the optimal policy such that $\pi^* = \arg\max_{\pi} q^\pi_G$ and $q^*_G := \max_{\pi}q_G^\pi = q^{\pi^*}_G$ be the optimal Q-value. Then, we can define the corresponding $N$-step regularized Bellman optimality operator.
\begin{definition}[Regularized Bellman optimality operator]
\label{def:bellman}
    For any $q\in\R^{|\St|\times|\A|}$, the $N$-step regularized Bellman optimality operator $(T_{G})^N$ is defined as
    \begin{align*}
        &(T_{G})^N  q (s_0,a_0)= \\
        &\EE \left[((1-\lambda)r(s_0,a_0)+(1-\gamma)\lambda G_\infty(s_0,a_0)) +\right.\\
        & +\gamma\max_{a_1}\EE\left[((1-\lambda) r(s_1,a_1)+ (1-\gamma)\lambda G_\infty(s_1,a_1)) \right.\\
        & +\gamma\max_{a_2}\EE\left[ ((1-\lambda)r(s_2,a_2)+(1-\gamma)\lambda G_\infty(s_2,a_2)) \right. \\
        & \left. \left. +\cdots + \gamma\max_{a_N}q(s_N,a_N) ] \right]  \vert s_0,a_0\right] ,
    \end{align*}
    where $G_\infty$ is the time-average growth rate of the best policy that can maximize $q^\pi_G$ over $\pi$.
\end{definition}
For $\lambda=1$, this operator maps any $q$ to a fixed point, $G_\infty$, the optimal time-average growth rate. 

Definition~\ref{def:bellman} is similar to \cite{carvalho2023multi}, and the operator satisfies the same properties. 

\begin{proposition}\label{prop:2}
The $N$-step regularized Bellman operator $(T_{G})^N$ has the following properties.
    \begin{itemize}
        \item[1.] The Q-value $q^{*}_{G}$ is a fixed point of the optimality operator $(T_{G})^N$. 
        \item[2.] Contraction: Let $\|\cdot\|_\infty$ be the infinity-norm for matrices. Then, for any $q_1,q_2\in\R^{|\St|\times|\A|}$,
        \begin{align*}
            \| (T_{G})^N q_1 - (T_{G})^N q_2  \|_\infty &\leq \gamma^N \| q_1 - q_2  \|_\infty.
        \end{align*}
    \end{itemize}
\end{proposition}

Our proposed objective function can be interpreted as a balance between the ensemble average and the time average, where $\lambda$ determines the weight given to the time average.

\subsubsection{Practical algorithm}
The properties of the Bellman operator explicitly support the development of a new algorithmic framework with an ideal time-average growth rate. In practice, when the ergodicity transformation 
is unknown, we must estimate it. A single-step return is insufficient to capture the full dynamics of the reward structure, which motivates our adoption of multi-step returns. Based on our previously proposed MGM~\eqref{equ:modified GM}, we introduce a novel algorithm that effectively estimates the time-average growth rate and incorporates it into the regularized MDP framework. 

We consider the data generated from a single trajectory. Under Assumption~\ref{Assp:1}, the Markov chain is stable after a mixing-time \cite{levin2017markov} and the visiting frequency of each state is stable. 
This implies that the state transitions resemble i.i.d.\ samples in the long run. 
However, the reward dynamics exhibit path dependence in non-ergodic settings. Therefore, the training data should cover a sufficiently long time horizon to capture the estimated growth rate.

To simplify notation, we define $ \mathcal{O}^{t+N-1}_{t} := ( (s_{t},a_{t},r_{t}),..., (s_{t+N-1},a_{t+N-1},r_{t+N-1}) )$ to denote the sequence of samples collected over one $N$-sliding window. Additionally, we introduce a dynamic awareness parameter $e\in\{0,1\}$ to indicate the mode of window progression. To be precise, if $e=0$, the sliding window moves in overlapping increments by discarding the first tuple of the previous window and appending a new tuple generated by the updated policy. Formally, the data combination $\mathcal{O}^{t+N}_{t+1}$ will be used in the next iteration. Conversely, if $e=1$, the window advances in non-overlapping segments and all data within a window is generated by the same policy. Then, $\mathcal{O}^{t+2N-1}_{t+N}$ will be used in the next iteration. 

We summarize our approach in Algorithm~\ref{alg:1}.

\begin{algorithm}
\caption{Regularized multi-step Q-learning with MGM. }
\label{alg:1}
\begin{algorithmic}
\STATE Initialize regularized Q-function $q_0$, policy $\pi_0$, learning rate $\alpha$, dynamic awareness $e$, and growth rate estimator $\hat{G}$ with size $N$. 
\FOR{$ t = 0,1,2,..., T$}    
    \IF{$e=0$}
    \STATE Sample $\mathcal{O}_t = \mathcal{O}^{t+N-1}_{t}$
    \ELSE 
    \STATE Sample $\mathcal{O}_t = \mathcal{O}^{(t+1)N-1}_{tN}$
    \ENDIF
    \STATE Compute $\delta^{(N)}_t$ and $\hat{G}_t$ based on $\mathcal{O}_t$
    \STATE $q_{t+1}(s_0,a_0)\gets (1-\alpha) q_t(s_{0},a_{0}) + \alpha(\Delta_t + \max_{a'} q_t(s_{N-1}),a')$, where $\Delta_t= (1-\lambda)\delta^{(N)}_t + \lambda(1-\gamma^N)\hat{G}_t$
    \STATE $\pi_{t+1}(s) \gets \arg\max_{a}q_{t+1}(s,a) \quad \forall s.$
    
\ENDFOR
\RETURN $\pi_\mathrm{T}$
\end{algorithmic}
\end{algorithm}

\section{EVALUATION} 
\begin{figure}
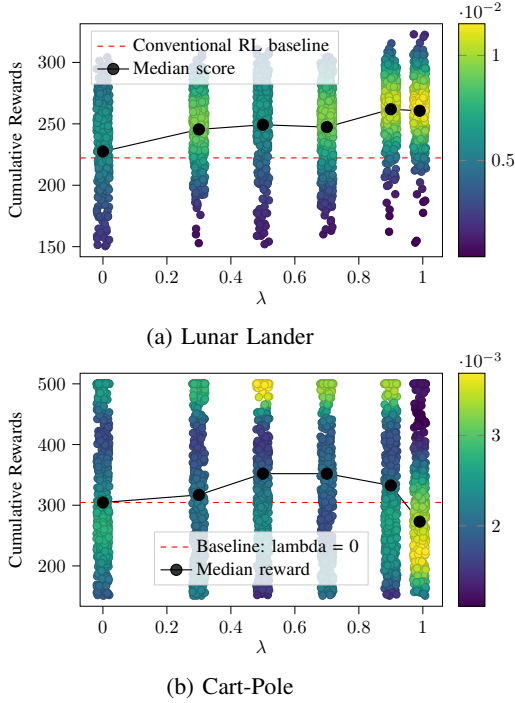

    \centering 
    \begin{subfigure}[b]{0.7\columnwidth}
        \centering
        \scalebox{0.7}{%
        \input{tikz/Lunar/Lunar_2step_median}
    }
        \caption{Lunar Lander}
    \end{subfigure}    
    \vspace{0.5em} 
    \begin{subfigure}[b]{0.7\columnwidth}
        \centering
        \scalebox{0.7}{%
        \input{tikz/CartPole/Cartpole_5step_median0}
    }
        \caption{Cart-Pole}
    \end{subfigure}
    \caption{Cumulative reward averaged over 100 episodes for (a) Lunar Lander and (b) Cart-Pole under varying $\lambda$. \capt{The color gradient bars indicate the density of the cumulative rewards. The red dashed lines represent the median value of standard multi-step Q-learning. In both environments, our algorithm learns higher performing policies for most choices of $\lambda$.}
     }
     \label{fig:1}
\end{figure}

In this section, we evaluate Algorithm~\ref{alg:1} in two simulation environments from \cite{brockman2016openai}: the Lunar Lander and the Cart-Pole. 
As our method modifies the multi-step return, we compare it with a conventional multi-step Q-learning baseline, which corresponds to our method with $\lambda=0$ and $e=0$, shown as the red dashed line in Figure \ref{fig:1}. For fairness, the baseline and our method share the same multi-step size $N$ and hyperparameters in each environment, varying only $\lambda$.\footnote{Code for all experiments is available at \href{https://github.com/Xinyi-Sheng18/CDC-2025-beyond-expectation}{https://github.com/Xinyi-Sheng18/CDC-2025-beyond-expectation}, including a coin toss example \cite{baumann2023reinforcement}. The Lunar Lander DQN is adapted from \href{https://www.katnoria.com/nb_dqn_lunar/}{https://www.katnoria.com/nb\_dqn\_lunar/}.}

In Lunar Lander, we use the multi-step deep Q-network (DQN) algorithm\cite{mnih2013playing} . We set $N=2$ and $e=1$ (without overlap) during training, as there is a specific reward dynamic that considers energy consumption, angle, and other factors. The testing results across all $\lambda$ show that our method with non-zero $\lambda$ outperforms traditional multi-step Q-learning. For the Cart-Pole, we discretise the state space and set $N=5$. The reward at each step is $+1$ if the pole is balanced, and $-1$ if it is dropped. Given the relatively ergodic reward dynamics, we set $e=0$ (with overlap) to better capture the reward variation. The results demonstrate that our method can achieve better performance with most $\lambda$ values. 

Compared with traditional methods, our method can pull up the median cumulative reward in both environments, and most of the cumulative rewards are concentrated at higher values. In fact, in addition to the median value, the mean value was also improved in our experiments. 

\section{CONCLUSION}
We developed an integrated framework based on regularized MDPs and multi-step Q-learning. By combining a sliding window mechanism with a modified geometric mean, we estimate the time-average of the reward dynamics. 
We then proposed a practical algorithm incorporating the modified geometric mean and multi-step Q-learning, which, when dynamic awareness $e=0$ and $\lambda=0$, defaults to multi-step Q-learning. 
The algorithm enables us to effectively balance the ensemble average and time-average, addressing the challenges posed by non-ergodic reward dynamics. Through the lens of reward growth for each individual, the algorithm can not only enhance the robustness of the learned policy but also improve the performance of the ensemble in complex environments. In future work, we plan to extend this approach to continuous spaces. 






\section*{APPENDIX}
\subsection{Proof of Proposition~\ref{prop:1}}
Consider the dynamics in discrete time. Let $R(t), \Delta R(t) $ denote the cumulative reward dynamic and its increment, respectively. Then $R(t+\Delta t) = R(t)+\Delta R(t)$. Let $r_i(t)$ denote the incremental reward dynamic for each state $i$ following a GBM with parameters $\mu_i$ and $\sigma_i$, that is $ \Delta R(t) = R(t)( \mu_i \Delta t +  \sigma_i dW_t ) $, where $ W_t$ is a Wiener process.

Starting from $R(t_0)$, let us consider a time interval with size $\delta$ and partition it into multiple segments with size $t_1,t_2,\cdots,t_n$ based on state transitions. We introduce $t_k,k=1,\cdots,n$ as the duration during which the system remains in a given state and follows its corresponding GBM until a transition (to the next state) occurs. Then, we have
\begin{align*}
    \ln{\frac{R(t_0+\delta)}{R(t_0)}} &= \sum_{k=0}^{n-1} \ln{\frac{R(t_0+\cdots+ t_{k+1})}{R(t_0+\cdots+t_k)}}\\
    &= \sum_{k=0}^{n-1} \left( (\mu_{k}-\frac{\sigma_k^2}{2})t_k + \sigma_k W_{t_k} \right)\\
    &=\sum_{k=1}^{n}(\mu_{k}-\frac{\sigma_k^2}{2})t_k + \sum_{k=1}^{n}\sigma_k W_{t_k}.
\end{align*}
The first linear term can be reconstructed by the stationary distribution for each state, since if $\delta$ is large enough, the frequency with which each state $i$ is visited converges to its stationary distribution $d^\pi_i$. Then, there exists
\begin{align*}
    \sum_{k=1}^{n}(\mu_{k}-\frac{\sigma_k^2}{2})t_k = \delta\sum_{i=1}^K d^\pi_i (\mu_i - \frac{\sigma_i^2}{2}).
\end{align*}
The second term is a linear combination of independent normal distributions $ W_{t_k}\sim \mathcal{N}(0,t_k) $:
\begin{align*}
    \sum_{k=1}^{n}\sigma_k W_{t_k}\sim \mathcal{N}(0,\sum_{k=1}^n \sigma_k^2 t_k).
\end{align*}
Then, we have
\begin{align*}
    \ln{\frac{R(t_0+\delta)}{R(t_0)}} \sim \mathcal{N}\left(\delta\sum_{i=1}^K d^\pi_i (\mu_i - \frac{\sigma_i^2}{2}), \sum_{k=1}^n \sigma_k^2 t_k) \right).
\end{align*}
Following \cite{peters2022ergodicity}, the time-average growth rate is the limit of
\begin{align*}
    \frac{1}{\delta}\ln{\frac{R(t_0+\delta)}{R(t_0)}} \sim \mathcal{N}\left(\sum_{i=1}^K d^\pi_i (\mu_i - \frac{\sigma_i^2}{2}), \frac{\sum_{k=1}^n \sigma_k^2 t_k) }{\delta^2}\right).
\end{align*}
The variance term will converge to $0$ when $\delta\to \infty$ as $\delta = \sum_{k}^n t_k$ and $\sigma_k$ are constants. Taking the limit $\delta\to \infty$ for both sides results in a fixed time-average growth rate $\sum_{i=1}^K d^\pi_i (\mu_i - \frac{\sigma_i^2}{2})$. 

\subsection{Proof for Proposition \ref{Assp:2}}
1.\ We define a Bellman operator as
\begin{align*}
    (T^{\pi}_{G})^N q(s_0,a_0)=& \EE^\pi \left[ \sum^{N-1}_{t=0}\gamma^t (1-\lambda)r_t + (1-\gamma^N)\lambda G_\infty\right. \\
        &  \left. +\gamma^N q(S_{N},A_{N}) | s_0,a_0\right].
\end{align*}
The fixed point for this Bellman operator is $q^{\pi}_{G}$, obtained by reshaping $\lambda G^\pi_\infty=(1-\gamma^N)G^\pi_\infty + \gamma^N G^\pi_\infty$. We show the fixed point for the Bellman optimality operator. For any $\pi$, 
\begin{align*}
    &q^{\pi}_{G}(s_0,a_0) = \EE^\pi \left[ \sum_{t=0}^\infty \gamma^t (1-\lambda)r_t + \lambda G^\pi_\infty \vert s_0 ,a_0\right]\\
    =&\EE^\pi\left[ \sum_{t=0}^{\infty}\gamma^t \left( (1-\lambda)r_t + (1-\gamma)\lambda G_\infty^\pi (s_t,a_t)  \right) |s_0,a_0 \right]  \\ 
    =& \EE^\pi \left[ \sum_{t=0}^{N-1} \gamma^t \left((1-\lambda)r_t + (1-\gamma)\lambda G^\pi_\infty(s_t,a_t)\right) \right.\\
    & + \gamma^N q^{\pi}_{G} (s_N,a_N)\left.  \vert s_0 ,a_0\right]\\
    \leq& \EE \left[ \left((1-\lambda)r_0+(1-\gamma)\lambda G^\pi_\infty(s_0,a_0)\right) \right.\\
    &+\gamma\max_{a_1}\EE\left[\left((1-\lambda)r(s_1,a_1) + (1-\gamma)\lambda G^\pi_\infty(s_1,a_1) \right)\right.\\
    & +\gamma\max_{a_2}\EE\left[\left((1-\lambda)r(s_2,a_2) + (1-\gamma)\lambda G^\pi_\infty(s_2,a_2)\right) \right. \\
    & \left. \left. +\cdots + \gamma \max_{a_N}q^{\pi}_{G}(s_N,a_N) ] \right] \vert s_0,a_0\right].
\end{align*}
Since $\pi$ is arbitrary, $q^{*}_{G} \leq (T_G)^N q^{*}_{G}$. For the other inequality, we choose a policy $\pi$ such that at $t=N$, it chooses an action $a_N$ such that $ q^{\pi}_{G}(s_N,a_N)\geq q^{*}_{G}(s_N,a_N)-\epsilon $, and at $t={1,2,\cdots,N-1}$, it selects 
\begin{align*}
    &a_t=\arg\max_{a_t}\{\EE [((1-\lambda)r(s_t,a_t)+(1-\gamma)\lambda G_\infty(s_t,a_t))\\
    &\gamma\max _{a_{t+1}}\EE[((1-\lambda)r(s_{t+1},a_{t+1})+(1-\gamma)\lambda G_\infty(s_{t+1},a_{t+1}))\\
    &+\cdots+\gamma\max_{a_N}q^*_G (s_{N},a_{N}) ]] \}.
\end{align*}
Hence, for any $(s,a)$ there is
\begin{align*}
     q^{\pi}_{G} (s,a) \geq (T_G)^N q^{*}_{G}(s,a) - \gamma^N\epsilon.
\end{align*}
Together, we have $$(T_G)^N q^{*}_{G} - \gamma^N\epsilon \leq q^{\pi}_{G} \leq  (T_G)^N q^{*}_{G}. $$
Since $\epsilon$ is arbitrary, we get the equality $ q^{\pi}_{G} =  (T_G)^N q^{*}_{G} $ . 

2.\ The result is established by induction. Firstly, consider $N=1$. Then,
\begin{align*}
     &\| (T_{G}) q_1 - (T_{G}) q_2 \|_\infty \\
     =& \max_{s,a}| \EE[(1-\gamma)\lambda G_\infty + (1-\lambda)r(s,a)+ \gamma\max_{a_1}q_1(s_1,a_1)] \\
      &-  \EE[(1-\gamma)\lambda G_\infty + (1-\lambda)r(s,a)+ \gamma\max_{a_1}q_2(s_1,a_1)] |\\
      =& \gamma \max_{s,a}| \EE[ \max_{a_1}q_1(s_1,a_1) - \max_{a_1}q_2(s_1,a_1)] |\\
      \leq&  \gamma \max_{s,a}\EE[ |\max_{a_1}q_1(s_1,a_1) - \max_{a_1}q_2(s_1,a_1)|]\\
      \leq& \gamma \max_{s,a}\EE[  \max_{a_1} | q_1(s_1,a_1)-q_2(s_1,a_1) |  ].
\end{align*}
The first inequality is due to Jensen's inequality. As the expectation of a function is less than or equal to its maximum,  
\begin{align*}
    &\EE[  \max_{a_1} | q_1(s_1,a_1)-q_2(s_1,a_1) |  ]\\
    \leq& \max_{s_1,a_1} | q_1(s_1,a_1)-q_2(s_1,a_1) |.
\end{align*}
With this, we have
\begin{align*}
    \| (T_{G}) q_1 - (T_{G}) q_2 \|_\infty
    \leq &\gamma\max_{s_1,a_1} | q_1(s_1,a_1)-q_2(s_1,a_1) |\\
    =& \gamma \|q_1 -q_2\|_\infty.
\end{align*}
Now, making an induction step for $ T_{G}^{n+1} = T_{G}T_{G}^{n} $, we have
\begin{align*}
     \| T_{G}(T_{G}^{n} q_1) - T_{G}(T_{G}^{n} q_2) \|_\infty 
     \leq \gamma\| T_{G}^{n} q_1 - T_{G}^{n} q_2 \|_\infty.
\end{align*}
Therefore, $\| T_{G}^{N} q_1 - T_{G}^{N} q_2 \|_\infty \leq \gamma^N \|q_1 -q_2\|_\infty$.





\bibliographystyle{IEEEtran}
\bibliography{IEEEabrv,mybibfile}

\end{document}